\documentclass[lettersize,journal]{IEEEtran}
\usepackage{amsmath,amsfonts}
\usepackage{algorithmic}
\usepackage{algorithm}
\usepackage{array}
\usepackage[caption=false,font=normalsize,labelfont=sf,textfont=sf]{subfig}
\usepackage{textcomp}
\usepackage{stfloats}
\usepackage{url}
\usepackage{verbatim}
\usepackage{graphicx}
\usepackage{colortbl}
\usepackage{cite}
\usepackage{xcolor}
\usepackage{array}

\begin{document}

\title{A Non-Anatomical Graph Structure for isolated hand gesture separation in continuous gesture sequences}

\author{Razieh Rastgoo$^{1*}$, Kourosh Kiani$^{1}$,Sergio Escalera$^2$\\
$^1$Semnan University~~ $^2$Universitat de Barcelona and Computer Vision Center\\
rrastgoo@semnan.ac.ir, kourosh.kiani@semnan.ac.ir, sergio@maia.ub.es\\
*Corresponding Author}

\maketitle

\begin{abstract}
Continuous Hand Gesture Recognition (CHGR) has been extensively studied by researchers in the last few decades. Recently, one model has been presented to deal with the challenge of the boundary detection of isolated gestures in a continuous gesture video [17]. To enhance the model performance and also replace the handcrafted feature extractor in the presented model in \cite{17}, we propose a GCN model and combine it with the stacked Bi-LSTM and Attention modules to push the temporal information in the video stream. Considering the breakthroughs of GCN models for skeleton modality, we propose a two-layer GCN model to empower the 3D hand skeleton features. Finally, the class probabilities of each isolated gesture are fed to the post-processing module, borrowed from \cite{17}. Furthermore, we replace the anatomical graph structure with some non-anatomical graph structures. Due to the lack of a large dataset, including both the continuous gesture sequences and the corresponding isolated gestures, three public datasets in Dynamic Hand Gesture Recognition (DHGR), RKS-PERSIANSIGN, and ASLVID, are used for evaluation. Experimental results show the superiority of the proposed model in dealing with isolated gesture boundaries detection in continuous gesture sequences.
\end{abstract}


\section{Introduction}
\IEEEPARstart{H}{and} gestural communications are widely used in different types of communications in our daily life \cite{1}. Hand sign language, as the most grammatically structured category of these communications, is the preliminary language of the Deaf community \cite{2}. In addition to this, many other application areas use the gestural communications, such as motion analysis and human-computer interaction \cite{3}. Vision-based dynamic hand gesture recognition aims to recognize gesture labels from video inputs. This task is an active research area in computer graphics and human-computer interaction with a wide range of applications in VR/AR \cite{4}, healthcare \cite{5}, and robotics \cite{6}. There are two types of vision-based dynamic hand gesture recognition: Isolated and Continuous. Continuous hand gesture recognition is different from isolated gesture classification \cite{2} or gesture spotting \cite{7}, which is to detect the predefined hand gestures from a video stream and the supervision includes exact temporal locations for each gesture. The critical point in continuous hand gesture recognition is that each video stream of gesture sentence contains its ordered gloss labels but no time boundaries for each gloss. So, the main problem with continuous hand gesture recognition is learning the corresponding relations between the image time series and the sequences of glosses.

Recently, Deep Learning models have achieved breakthroughs in many tasks, such as Human Action Recognition \cite{8,9}, Hand Gesture Recognition \cite{10}, and Isolated Sign Language Recognition \cite{11,12,13,14,15,16,17,18,19}. However, continuous hand gesture recognition remains still challenging and non-trivial. More concretely, the recognition system needs to obtain spatio-temporal features from the weakly supervised unsegmented video stream. Since data includes the video sequences along with the gloss labels at the sentence level, a large amount of data samples is required to align the gestures and gloss labels correctly without overfitting. Furthermore, the boundaries of isolated gestures are not available in datasets. To tackle this challenge, recently, one model has been presented to solve the challenge of the boundary detection of isolated gestures in a continuous gesture video \cite{17}. The presented model in \cite{17} could efficiently detect the isolated gestures in a continuous video stream. However, some handcrafted features, obtained from the Singular Value Decomposition (SVD) method, have been used in that model. To improve the recognition accuracy and also use the fully automated features in that model, we propose a graph-based model, considering the recent breakthroughs of Graph Convolutional Network (GCN) in many research areas, especially gesture recognition \cite{20}. To this end, we combine GCN with the stacked Bi-LSTM and Attention modules to push the temporal information in the video stream. In addition to the aforementioned advantages, the GCN-based model decreases the computational complexity of the model that can be useful for real-time applications. 

The majority of the proposed models for CHGR have used deep Convolutional Neural Networks (CNNs) to obtain the features in each frame of the continuous video. After that, these features are fed to a temporal learning model, such as RNN, LSTM, or GRU. This combination of a CNN and RNN/LSTM/GRU is usually sensitive to the noisy background, occlusions, and different camera viewpoints. To tackle these challenges, another modality, the human body/hand skeleton, has been used by many researchers in other research areas \cite{1,7,9,14}. The more compact representation, better robustness against occlusion and viewpoint changes, and higher expressive capability in capturing features in both temporal and spatial domains are some of the advantages of the skeleton modality over the image/video modalities. A tailored way to represent the human skeleton is using graphs. To this end, the skeleton joints and bones are defined as graph nodes and edges, respectively. Here, a model is needed to extract features from graphs that GCN is the most popular and effective model in this way. Considering the aforementioned points and also the scope of this work, we propose a GCN-based model along with the Bi-LSTM-Attention block and also a post-processing algorithm \cite{17} for isolated hand gesture recognition in a continuous hand gesture video stream. The proposed non-anatomical graph aims to revolutionize the graph structure used in CHGR. This comes from this point that we do not necessarily need to use the anatomical structure of the hand skeleton. Results confirm that we need to rethink the graph structure used in human/hand skeleton-based graphs. 
Our contributions can be summarized as follows:
\begin{itemize}
    \item \textbf{Model}: We propose a GCN-based model as an efficient feature extractor (Graph embedding) along with the Bi-LSTM-Attention block for the isolated hand gestures recognition in a continuous hand gestures video stream.
    \item \textbf{Non-anatomical skeleton graph structure}: We propose a non-anatomical skeleton graph structure in the GCN model. However, the anatomical structure of the skeleton is usually used in Human Action and Sign Language Recognition.
    \item \textbf{Performance}: Providing further analyzes of the proposed model, the results on two large-scale datasets confirm the efficiency of the proposed graph structure. Our model obtains state-of-the-art results for isolated hand gesture recognition in the continuous gesture video.
\end{itemize}
The remaining paper is organized as follows. Related literature on deep models in CHGR is reviewed in section II. Details of the proposed methodology are described in section III. Results are presented and discussed in section IV. Finally, the work is discussed and concluded in section V.

\section{Literature review}
Here, we briefly review recent works in CHGR, Sign Language Recognition (SLR), and Human Action Recognition (HAR). 
Meng and Li proposed a multi-scale and dual sign language recognition Network (SLR-Net) using a Graph Convolutional Network (GCN). After extracting the skeleton data from the RGB video inputs, the skeleton data is used for Sign Language Recognition (SLR). The proposed model consists of three main parts: multi-scale attention network (MSA), multi-scale spatiotemporal attention network (MSSTA), and attention enhanced temporal convolution network (ATCN). These parts are used to learn the dependencies between long-distance vertices, the spatiotemporal features, and also the long temporal dependencies, respectively. Results on two datasets, CSL-500 and DEVISIGN-L, show that the proposed model outperforms state-of-the-art models in SLR \cite{21}. 

Vazquez-Enrıquez et al. proposed a graph-based model for skeleton-based Isolated Sign Language Recognition (ISLR). They benefit from the advantages of the multi-scale spatial-temporal graph convolution operator, MSG3D, to use the semantic connectivity among non-neighbor nodes of the graph in a flexible temporal scale. Results on the AUTSL dataset show promising results in ISLR \cite{22}. 

Degardin et al. introduce a model, entitled REasoning Graph convolutional networks IN Human Action recognition (REGINA), to include handcrafted features to the GCN model and facilitate the learning process. To this end, a handcrafted Self-Similarity Matrix (SSM) is applied to the temporal graph convolution part, aimed to enhance the global connectivity across the temporal axis. Results on the NTU RGB+D dataset show competitive results concerning the state-of-the-art models in skeleton-based action recognition \cite{23}.

Duhme et al. proposed a graph-based model for action recognition from various sensor data modalities. Different modalities are fused on two-dimensionality levels: channel dimension and spatial dimension. Results on two publicly available datasets, UTD-MHAD and MMACT, demonstrate that the proposed model obtains a relative improvement margin of up to 12.37\%\ (F1-Measure) with the fusion of skeleton estimates and accelerometer measurements \cite{24}.

Li et al. proposed an encoder-decoder structure to obtain action-specific latent dependencies from actions. To this end, they include higher order dependencies of the skeletal data in the graph structure. Stacking some instances of the proposed graph, the proposed model aims to learn both spatial and temporal features for action recognition. Results on the NTURGB+D and Kinetics datasets show that the proposed model outperforms the state-of-the-art models in action recognition \cite{25}. 

Rastgoo et al. proposed a two-stage model, including a combination of CNN, SVD, and LSTM. After training with the isolated signs, a post-processing algorithm is applied to the Softmax outputs obtained from the first part of the model in order to separate the isolated signs in the continuous signs. Results of the continuous sign videos, created using two public datasets in Isolated Sign Language Recognition (ISLR), RKS-PERSIANSIGN and ASLVID, confirm the efficiency of the proposed model in dealing with isolated sign boundaries detection \cite{17}.

\section{Proposed model}
Here, we describe the details of the proposed model, as Fig. 1 shows.

\begin{figure*}[!t]
\centering
{\includegraphics[width=0.8\linewidth]{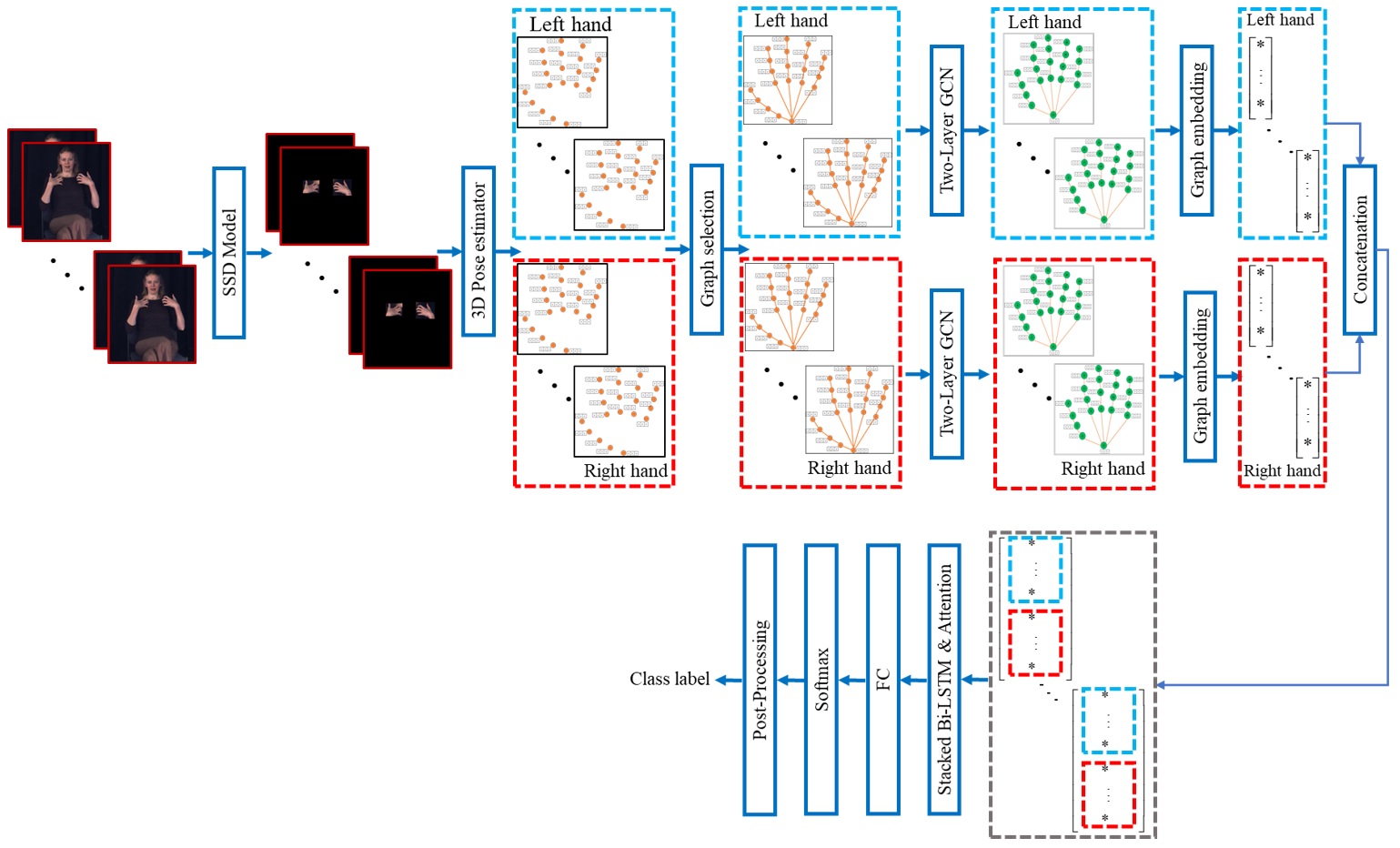}}
\caption{The proposed model.}
\label{Fig1}
\end{figure*}

Recently, one model has been presented to solve the challenge of the boundary detection of isolated gestures in a continuous gesture video \cite{17}. To enhance the model performance and also replace the handcrafted feature extractor in the presented model in \cite{17}, we propose a GCN model and combine it with the stacked Bi-LSTM and Attention modules to push the temporal information in the video stream. Considering the breakthroughs of GCN models for skeleton modality, we propose a two-layer GCN model to empower the 3D hand skeleton features. The obtained features are fed to the stacked Bi-LSTM and attention module. The class probabilities of each isolated gesture are fed to the post-processing module, borrowed from \cite{17}. However, the post-processing module is only used in the test phase. More concretely, the proposed model contains the following steps:
\begin{itemize}
    \item \textbf{Input}: We have two input types, fed to the model frame-by-frame, as follows: 
    \begin{itemize}
        \item \textbf{Isolated gesture videos}: This input contains the videos of isolated gestures used for model training.
        \item \textbf{Continuous gesture video}: This input contains the concatenated isolated gesture videos without any pre-processing, which is fed to the trained model. The model uses this input for the separation of the isolated gestures in the continuous gesture videos. 
    \end{itemize}
    \item \textbf{Hand detection}: To improve the hand detection accuracy in the model, we trained the Single Shot Detector (SSD) using some online gesture dictionaries. More details of the training mechanism can be found in \cite{12}.
    \item \textbf{3D hand pose estimation}: The OpenPose model, as the state-of-the-art model for pose estimation, is used for 3D hand pose estimation. 21 3D hand keypoints are obtained and used in the proposed model. For simplicity, we normalize these coordinates in [0, 96] interval.
    \item \textbf{Graph selection}: After analyzing the hand skeleton with anatomical and non-anatomical structures, we are encouraged to replace the anatomical graph structure with some non-anatomical graph structures. We have experimentally checked many different graph structures. However, only three structures have been shown and reported in this work (See Fig. 2). Our study show that we do not necessarily need to use the anatomical structure of the hand skeleton. Each proposed graph includes 21 nodes that the estimated 3D hand keypoints are considered as the node property/message. 
    \item \textbf{Two-Layer GCN}: We use a GCN-based model as an efficient feature extractor to enrich the 3D hand features of both hands. Both the input and output of the GCN are similar graphs. The only difference between them is the node embeddings. The GCN uses the neighbor nodes in a graph to empower the node features. So, the output graph obtained from the GCN is a graph with richer features for each node. For each hand, we use a two-layer GCN, separately.
    \item \textbf{Graph embedding}: The property/message of all nodes in the graph are flattened to a vector called graph embedding. This process is separately performed for each hand.
    \item \textbf{Concatenation}: The graph embedding corresponding to the left and right hands are concatenated frame-by-frame and fed to the next part of the model.
   \item \textbf{Stacked Bi-LSTM and Attention}: To obtain the temporal information in the video stream, a stacked Bi-LSTM and Attention module is used. The output of this step is a vector that is fed to the next part of the model.
   \item \textbf{Fully Connected (FC)}: We use the FC layers to enhance the model learning. 
   \item \textbf{Softmax}: The class probabilities for the input video are obtained in the Softmax layer.
   \item \textbf{Post-Processing}: We borrow this step from our previous work \cite{17}. Here, we apply a post-processing algorithm to the Softmax outputs to recognize the isolated gestures in a continuous gesture sequence. This step is only used for continuous gesture videos. More details of this step can be found in \cite{17}.
\end{itemize}

\begin{figure*}[!t]
\centering
{\includegraphics[width=0.8\linewidth]{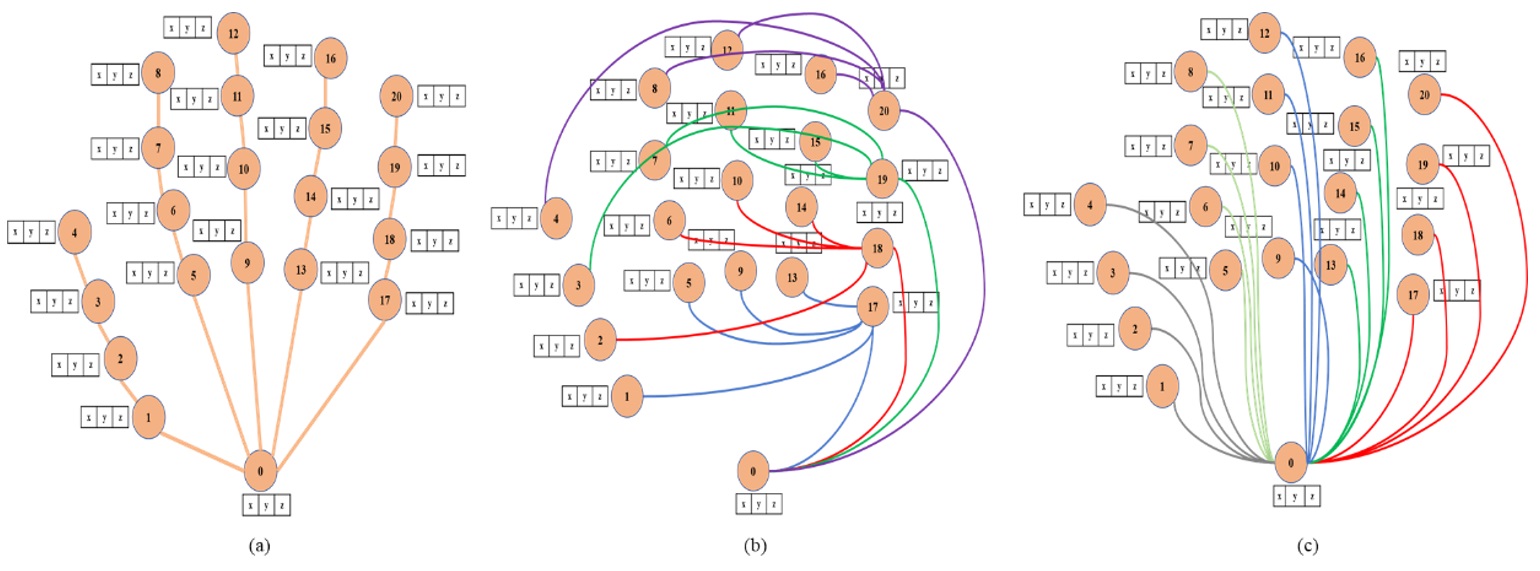}}
\caption{Three graph structures used in the proposed model: (a) Anatomical hand skeleton graph, (b), (c) Non-Anatomical graph structures.}
\label{Fig1}
\end{figure*}

\section{Anatomical vs. Non-Anatomical Graph structure}
As Fig. 3-a shows, in the anatomical structure, the hand is separated into two parts: the fingers and the palm. Considering the joints of each finger, four keypoints are considered in each finger. Only one keypoint is assigned to the hand palm. After investigating the hand anatomy and also the patterns of different hand gestures, we observed that there are many alternatives for anatomical graph structures, which are extensively used in literature. We experimentally checked and analyzed many of them. However, we only present three of them in the paper, including the anatomical graph structure and two best non-anatomical graph structures with the highest accuracy. All of the proposed structures consider the natural hand, including five fingers that each finger has four joint connections. To model the hand in this structure, four keypoints are considered on each finger, as Fig. 3-b shows. Another keypoint is put on the hand palm. So, we will have 21 keypoints on each hand in all proposed hand structures. The difference between the proposed hand structures is in the connections between these keypoints. Details of these structures are described as follows:
\begin{itemize}
    \item \textbf{Anatomical structure}: This structure is based on the natural anatomy of the hand, which is extensively used in literature. In this structure, a keypoint, put on the hand palm, is connected to all keypoints near this keypoint. The keypoints put on each finger are separately connected to each other. Here, we have 20 connections between 21 hand keypoints. The pattern of this structure is shown in Fig. 3-c.
    \item \textbf{Non-Anatomical structure with one finger plus one point references}: In this structure, four keypoints, put on one finger, and also one keypoint corresponding to the hand palm, are considered as reference points. The connections start from the reference points on a finger to the other aligned hand keypoints. Finally, the palm keypoint is connected to all keypoints on the reference finger. As Fig. 2-b, there are 20 connections in this structure. The intuition behind selecting all keypoints on a finger plus one palm keypoint as the reference points is the smooth movement of this finger and also the palm keypoint in different dynamic gestures. So, we referred all connections to this finger.
    \item \textbf{Non-Anatomical structure with one reference point}: Similar to the anatomical structure, we consider the palm keypoint as a reference point. However, the connections are referenced from this point to the other keypoints. As Fig. 2-c shows, we have 20 connections from the palm keypoint to the other 20 hand keypoints. The intuition behind the reference point is that the palm keypoint has a smooth movement in different gestures. So, we can consider it as a gravity center to handle the other keypoints.
\end{itemize}

\begin{figure*}[!t]
\centering
{\includegraphics[width=0.7\linewidth]{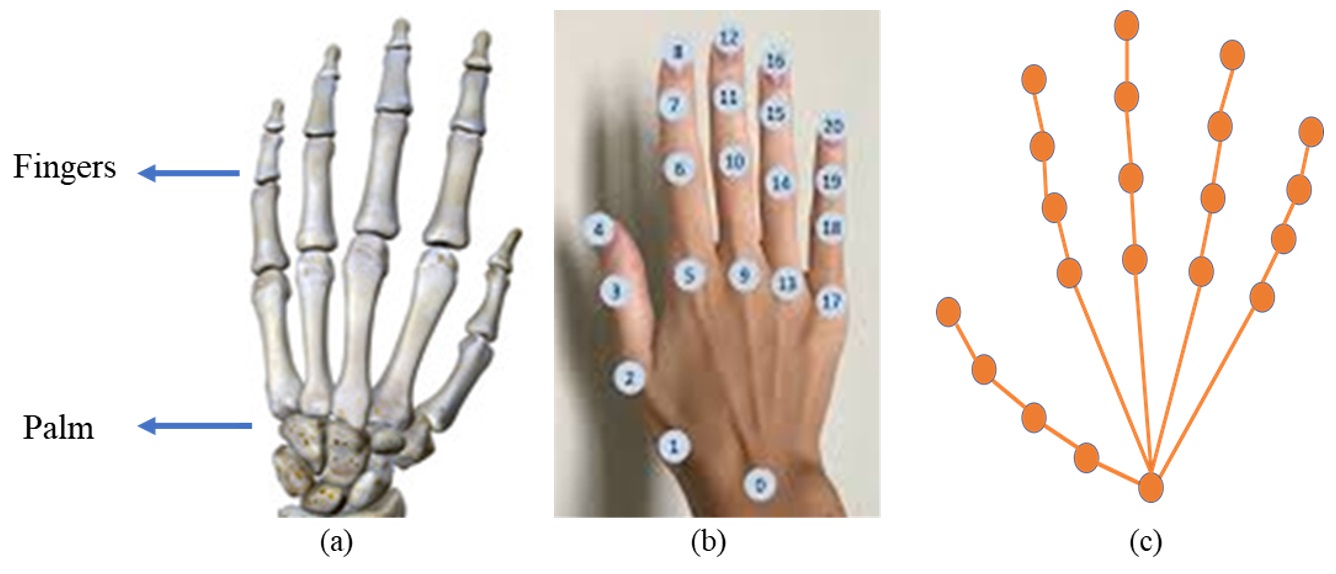}}
\caption{Three perspectives of the hand: (a) Anatomical, (b) Hand keypoints, (c) Hand skeleton.}
\label{Fig1}
\end{figure*}

\section{Results}
In this section, we present details of the datasets and results. First of all, the implementation details are described. After that, two datasets, used for evaluation, are briefly introduced. Finally, the details of the ablation analysis of the proposed graphs and also the comparison with state-of-the-art models are discussed.

\subsection{Implementation details}
Our evaluation has been done on an Intel(R) Xeon(R) CPU E5-2699 (2 processors) with 90GB RAM with Microsoft Windows 10 operating system and Python software with NVIDIA Tesla K80 GPU. The PyTorch library has been used for model implementation. The parameters used in the implementation are shown in TABLE I. 

\begin{table}[h]
\caption{Details of the parameters used in the proposed model.}
\centering
\begin{tabular}{|p{2.4cm}|p{0.6cm}|p{3.2cm}|p{0.6cm}|}
\hline
Parameter & Value & Parameter & Value\\
\hline
Weight decay & 1e-4 & Epoch numbers (Early stopping) & 200\\
Learning rate & 0.005 & Number of frames per video sample & 50\\
Batch size & 50 & Processing way & GPU\\
Keypoint dimension & 21x3 & Number of GCN layer & 2\\
Optimizer & Adam & Dataset split ration for test data &  20\%\\\
GCN layer numbers & 2 & Aggregation type & ‘Sum’\\
\hline
\end{tabular}
\end{table}

\subsection{Datasets}
The available datasets in CHGR do not contain the pair of continuous gesture videos and the corresponding isolated gestures in each video. So, to cope with this challenge, two datasets in IHGR are utilized for evaluation, RKS-PERSIANSIGN \cite{14} and ASLVID \cite{26}. In RKS-PERSIANSIGN, 10 contributors performed 100 Persian gestures 10 times in a simple background with a maximum distance of 1.5 meters between the contributor and camera. So, there are 10000 RGB videos in this dataset. The ASLVID dataset includes some annotated American isolated gestures. Due to the different sample numbers in each class label, 100 gestures, including at least seven video samples, are selected and used. Using the pre-processing, all video samples have equally frame numbers during the training phase. Whereas, we do not perform any pre-processing on the test data and use them with different frame numbers.

\subsection{Ablation analysis}
Here, we perform an ablation analysis on two datasets used for evaluation (TABLE II). In the first step, we used the GCN along with the LSTM Network. To improve the model performance, different GCN, LSTM, and FC layer numbers have been experimented and analyzed. The analysis showed that the model performance was improved using higher FC layers. However, the improvement trend was stopped with four FC layers. So, we had the highest performance using two, one, and four layers for GCN, LSTM, and FC, respectively. After that, we substituted the LSTM with Bi-LSTM Network. Unlike the standard LSTM, the input flows in Bi-LSTM are in both directions. Furthermore, it is capable of utilizing information from both sides. Results confirmed the efficiency of the Bi-LSTM Network. We also analyzed the impact of different layer numbers for the Bi-LSTM Network. As the trend of this table shows, the highest accuracy is obtained using three Bi-LSTM layers (Last row of TABLE II). However, we only reported the results of three graph structures with GCN, Bi-LSTM, and FC combination. 

\begin{table*}[h]
\caption{Ablation analysis of the proposed model using different graph topology and model architecture on two datasets.}
\centering
\begin{tabular}{p{2cm}p{5.5cm}p{3cm}p{2cm}p{0.5cm}p{0.5cm}p{0.5cm}} 
Graph Topology & Model & FC & LSTM Hidden & Epoch & Accuracy &\\
 &  &  &  &  & 	RKS	& ASL\\
\hline
\hline
(a) & GCN-BiLSTM-Attention & 4 (512-256-128-100) & 400-400 & 10000 & 87.40 & 86.00\\
(a) & GCN-BiLSTM-BiLSTM -BiLSTM-Attention & 4 (512-256-128-100) & 400-400 &10000 & 87.80 & 87.00\\
(b) & GCN-BiLSTM-BiLSTM-Attention & 4 (512-256-128-100) & 400-400 & 10000 & 99.80 & 94.80\\
(b) & GCN-BiLSTM-BiLSTM -BiLSTM-Attention & 4 (512-256-128-100) & 400-400 & 10000 & 99.85 & 95.50\\
(c) & GCN-BiLSTM-BiLSTM-Attention & 4 (512-256-128-100) & 400-400 & 10000 & 99.85 & 95.25\\
(c) & GCN-BiLSTM-BiLSTM -BiLSTM-Attention & 4 (512-256-128-100) & 400-400 & 10000 & 99.90 & 96.00\\
\hline
\hline
\end{tabular}
\end{table*}

\subsection{Comparison with state-of-the-art}
Here, we compare the current model in this work with our previous work \cite{17}, which has been proposed to isolated gesture separation in the continuous gesture video. As TABLE III and TABLE IV show, the proposed model in this work has a better performance. As the number of false recognition and also the average of the recognized Softmax outputs on the RKS and ASL datasets show, there are still some false recognition coming from the similarities between the gestures in the datasets. In the proposed model, the average of recognized Softmax outputs on RKS-PERSIANSIGN and ASLVID are 0.99 and 0.68, respectively. 

\begin{table}[h]
\caption{Comparison with the state-of-the-art model.}
\centering
\begin{tabular}{p{2cm}p{2cm}p{1cm}}
\hline
Model & \multicolumn{2}{c}{Number of false recognition}\\
\hline
 & \multicolumn{2}{c}{RKS}{ASL}\\
\cite{17} & \multicolumn{2}{c}{24}{12}\\
Proposed model & \multicolumn{2}{c}{9}{7}\\
\hline
\end{tabular}
\end{table}

\begin{table}[h]
\caption{The average of recognized Softmax outputs on the RKS and ASL datasets.}
\centering
\begin{tabular}{p{2cm}p{2cm}p{1cm}}
\hline
Model & \multicolumn{2}{c}{The average of recognized Softmax outputs}\\
\hline
 & \multicolumn{2}{c}{RKS}{ASL}\\
\cite{17} & \multicolumn{2}{c}{0.98}{0.59}\\
Proposed model & \multicolumn{2}{c}{0.99}{0.68}\\
\hline
\end{tabular}
\end{table}

\section{Discussion and conclusion}
We discuss the proposed model from three perspectives as follows:
\begin{itemize}
    \item \textbf{Model}: Considering the recently presented model that aims to solve the challenge of the boundary detection of the isolated gestures in a continuous gesture video \cite{17}, we proposed to enhance the model performance and also replace the handcrafted feature extractor in the presented model in \cite{17}. To this end, we proposed a GCN model and combined it with the stacked Bi-LSTM and Attention module to push the temporal information in the video stream. Relying on the capabilities of GCN, Bi-LSTM, Attention, and also the post-processing algorithm borrowed from the \cite{17}, the proposed model improved the accuracy of boundaries detection of isolated gestures in continuous gesture video. Different ablation analysis has been performed on different parts of the model to show the necessity of each part. Our analysis confirmed the efficiency of the Bi-LSTM Network vs. the LSTM Network due to the capability of utilizing information from both sides. However, we had an increasing trend in model performance by adding more Bi-LSTM Network only from one to three layers number. After that, model performance decayed. As TABLE II shows, the highest accuracy is obtained using the third graph topology combined with three Bi-LSTM and an Attention module.
    \item \textbf{Graph structure}: In this work, we proposed some non-anatomical graph structures, aiming to revolutionize the common graph structure used in CHGR. After investigating the hand anatomy and also the patterns of different hand gestures, we observed that there are many alternatives for anatomical graph structures, which are extensively used in literature. While all of the proposed non-anatomical and also the anatomical graph structures consider 21 keypoints on the hand, the difference between them is in the connections between these keypoints. Relying on the experimental results, the highest performance is obtained using a non-anatomical structure with one reference point. This structure considers the palm keypoint as a reference point that the connections are referenced from this point to the other keypoints.
    \item \textbf{Performance}: To enhance the model performance, we proposed a deep learning-based model, benefiting from the GCN capabilities. Since there is no any dataset containing both of the continuous gestures and the corresponding isolated gestures, we utilized the datasets in isolated hand gestures and concatenated them to make the continuous gesture videos. We compared the results of the proposed model with \cite{17}. While the results of both models confirm the efficiency of the proposed post-processing methodology for separation of isolated gestures in continuous gestures, the proposed model in this work has a higher recognition accuracy than the model in \cite{17}. Considering the GCN capabilities combined with Bi-LSTN and attention mechanism, the proposed model in this work has a more discriminative capability to truly classify the similar gestures. Furthermore, instead of using handcrafted features, such as singular values obtained from the Singular Value Decomposition (SVD) method, our model in this work uses the deep features obtained from the GCN model. Similar to the configuration of the post-processing methodology used in [17], we used a predefined threshold, 0.51, to accept or reject a recognized class in the current sliding window. Results on two datasets, as shown in TABLE V-VI, confirmed that if the model faces false recognition, the recognized Softmax outputs for all classes are lower than the predefined threshold. As these tables show, the proposed model obtains the average of recognized Softmax outputs of 0.68 and 0.99 on the RKS-PERSIANSIGN and ASLVID datasets, respectively. We have a higher recognition accuracy on the RKS-PERSIANSIGN dataset than the ASL due to higher video sample instances in each class. This comes from the fact that deep learning-based models generally have a better performance if they train with a large amount of data. Furthermore, some false recognition can be found in the similar gestures, such as 'Congratulation', 'Excuse', 'Upset', 'Blame', 'Fight', 'Competition'. For instance, 'Excuse' and 'Congratulation', 'Upset' and 'Blame', 'Fight' and 'Competition' gestures contain many similar frames. Thus, extending the gesture samples in these similar classes could lead to learning more powerful features and better representation of gesture categories. This also can decrease miss-classifications due to low inter-class variabilities. In future work, we aim to collect a dataset, including more realistic continuous gesture videos and the corresponding isolated gesture videos. Relying on this dataset, we can check the performance of the proposed model to use in a realistic scenario. Furthermore, we aim to apply our non-anatomical graph structure idea to Human Action Recognition.
\end{itemize}

\begin{table}[h]
\caption{Details of the recognition accuracy of the proposed post-processing algorithm on the ASLVID dataset. CSV: Concatenated Sign Video, ASORC: Avg. of Softmax Output of Recognized Class, GTWC: Ground Truth Word Class, SOGTWC: Softmax output of Ground truth Word class, RC: Recognized Class, SORC: Softmax output of Recognized class}
\centering
\begin{tabular}{p{0.3cm}p{0.7cm}p{0.6cm}p{0.9cm}p{0.6cm}p{0.6cm}} 
\hline
\hline
CSV	& ASORC	& GTWC & SOGTWC & RC & SORC \\
\hline
\hline
1 & 0.68 & \leavevmode\color{blue} 18 & \leavevmode\color{blue} 0.33 & \leavevmode\color{red} 8 & \leavevmode\color{red} 0.35\\
2 & 0.69 & \leavevmode\color{blue} 8 & \leavevmode\color{blue} 0.37 & \leavevmode\color{red} 18 & \leavevmode\color{red} 0.38\\
3 & 0.68 & \leavevmode\color{blue} 80 & \leavevmode\color{blue} 0.44 & \leavevmode\color{red} 50 & \leavevmode\color{red} 0.45\\
4 & 0.69 & \leavevmode\color{blue} 18 & \leavevmode\color{blue} 0.38 & \leavevmode\color{red} 8 & \leavevmode\color{red} 0.39\\
5 & 0.68 & \leavevmode\color{blue} 64 & \leavevmode\color{blue} 0.34 & \leavevmode\color{red} 51 &\leavevmode\color{red} 0.35\\
6 & 0.69 & \leavevmode\color{blue} 51 & \leavevmode\color{blue} 0.34 & \leavevmode\color{red} 64 & \leavevmode\color{red} 0.37\\
7 & 0.68 & \leavevmode\color{blue} 51 & \leavevmode\color{blue} 0.34 & \leavevmode\color{red} 64 & \leavevmode\color{red} 0.35\\
\hline
\hline
\end{tabular}
\end{table}

\begin{table*}[h]
\caption{Details of the recognition accuracy of the proposed post-processing algorithm on the RKS-PERSIANSIGN dataset. CSV: Concatenated Sign Video, ASORC: Avg of Softmax Output of Recognized Class, GTWC: Ground Truth Word Class, SOGTWC: Softmax output of Ground truth Word class, RC: Recognized Class, SORC: Softmax output of Recognized class}
\centering
\begin{tabular}{p{0.3cm}p{0.7cm}p{0.6cm}p{0.9cm}p{0.6cm}p{0.6cm}p{0.3cm}p{0.7cm}p{0.6cm}p{0.9cm}p{0.6cm}p{0.7cm}p{0.3cm}p{0.9cm}p{0.6cm}p{0.9cm}p{0.3cm}p{0.6cm}}
\hline
CSV	& ASORC	& GTWC & SOGTWC & RC & SORC & CSV & ASORC & GTWC & SOGTWC & RC & SORC & CSV	& ASORC	& GTWC & SOGTWC & RC & SORC\\
\hline
1 & 0.98 & \leavevmode\color{blue} 17 & \leavevmode\color{blue} 0.45 & \leavevmode\color{red} 19 & \leavevmode\color{red} 0.46 & 34 & 0.97 & \leavevmode\color{blue} 86 & \leavevmode\color{blue} 0.44 & \leavevmode\color{red} 66 & \leavevmode\color{red} 0.49 & 68 & 0.98 & \leavevmode\color{blue} 63 & \leavevmode\color{blue} 0.48 & \leavevmode\color{red} 45 & \leavevmode\color{red} 0.49\\
2 & 0.98 & - & - & - & - & 35 & 0.99 & - & - & - & - & 69 & 0.99 & - & - & - & -\\
3 & 0.98 & - & - & - & - & 36 & 0.99 & - & - & - & - & 70 & 0.99 & - & - & - & -\\
4 & 0.99 & - & - & - & - & 37 & 0.99 & - & - & - & - & 71 & 0.99 & - & - & - & -\\
5 & 0.99 & - & - & - & - & 38 & 0.99 & - & - & - & - & 72 & 0.99 & - & - & - & -\\
6 & 0.99 & - & - & - & - & 39 & 0.99 & - & - & - & - & 73 & 0.99 & - & - & - & -\\
7 & 0.98 & - & - & - & - & 40 & 0.98 & - & - & - & - & 74 & 0.99 & - & - & - & -\\
8 & 0.99 & - & - & - & - & 41 & 0.99 & - & - & - & - & 75 & 0.99 & - & - & - & -\\
9 & 0.99 & - & - & - & - & 42 & 0.99 & - & - & - & - & 76 & 0.99 & - & - & - & -\\
10 & 0.98 & \leavevmode\color{blue} 19 & \leavevmode\color{blue} 0.46 & \leavevmode\color{red} 17 & \leavevmode\color{red} 0.47 & 43 & 0.99 & - & - & - & - & 77 & 0.97 & \leavevmode\color{blue} 63 & \leavevmode\color{blue} 0.47 & \leavevmode\color{red} 45 & \leavevmode\color{red} 0.48\\
11 & 0.99 & - & - & - & - & 44 & 0.98 & \leavevmode\color{blue} 45 & \leavevmode\color{blue} 0.45 & \leavevmode\color{red} 63 & \leavevmode\color{red} 0.46 & 78 & 0.99 & - & - & - & - \\
12 & 0.99 & - & - & - & - & 45 & 0.99 & - & - & - & - & 79 & 0.99 & - & - & - & -\\
13 & 0.99 & - & - & - & - & 46 & 0.99 & - & - & - & - & 80 & 0.99 & - & - & - & -\\
14 & 0.99 & - & - & - & - & 47 & 0.99 & - & - & - & - & 81 & 0.99 & - & - & - & -\\
15 & 0.99 & - & - & - & - & 48 & 0.99 & - & - & - & - & 82 & 0.99 & - & - & - & -\\
16 & 0.99 & - & - & - & - & 49 & 0.99 & - & - & - & - & 83 & 0.99 & - & - & - & -\\
17 & 0.99 & - & - & - & - & 50 & 0.99 & - & - & - & - & 84 & 0.99 & - & - & - & -\\
18 & 0.99 & - & - & - & - & 51 & 0.99 & - & - & - & - & 85 & 0.99 & - & - & - & -\\
19 & 0.99 & - & - & - & - & 52 & 0.99 & - & - & - & - & 86 & 0.99 & - & - & - & -\\
20 & 0.99 & - & - & - & - & 53 & 0.99 & - & - & - & - & 87 & 0.99 & - & - & - & -\\
21 & 0.99 & - & - & - & -  & 54 & 0.99 & - & - & - & - & 88 & 0.99 & - & - & - & -\\
22 & 0.99 & - & - & - & - & 55 & 0.99 & - & - & - & - & 89 & 0.99 & - & - & - & -\\
23 & 0.99 & - & - & - & - & 56 & 0.99 & - & - & - & - & 90 & 0.99 & - & - & - & -\\
24 & 0.99 & - & - & - & - & 57 & 0.99 & - & - & - & - & 91 & 0.99 & - & - & - & -\\
25 & 0.97 & \leavevmode\color{blue} 19 & \leavevmode\color{blue} 0.46 & \leavevmode\color{red} 17 & \leavevmode\color{red} 0.49 & 58 & 0.99 & - & - & - & - & 92 & 0.99 & - & - & - & -\\
26 & 0.99 & - & - & - & - & 59 & 0.98 & \leavevmode\color{blue} 17 & \leavevmode\color{blue} 0.46 & \leavevmode\color{red} 19 & \leavevmode\color{red} 0.48 & 93 & 0.97 & \leavevmode\color{blue} 45 & \leavevmode\color{blue} 0.45 & \leavevmode\color{red} 63 & \leavevmode\color{red} 0.46 \\
27 & 0.99 & - & - & - & - & 60 & 0.99 & - & - & - & - & 94 & 0.99 & - & - & - & -\\
28 & 0.99 & - & - & - & - & 61 & 0.99 & - & - & - & - & 95 & 0.99 & - & - & - & -\\
29 & 0.99 & - & - & - & - & 62 & 0.99 & - & - & - & - & 96 & 0.99 & - & - & - & -\\
30 & 0.99 & - & - & - & - & 63 & 0.99 & - & - & - & - & 97 & 0.99 & - & - & - & -\\
31 & 0.99 & - & - & - & - & 64 & 0.99 & - & - & - & - & 98 & 0.99 & - & - & - & -\\
32 & 0.99 & - & - & - & - & 65 & 0.99 & - & - & - & - & 99 & 0.99 & - & - & - & -\\
 &  &  &  &  & & 66 & 0.99 & - & -  & 	-  &  -	 &  &  &  &  &  & \\		
\hline
\end{tabular}
\end{table*}

\newpage

\vspace{11pt}

\vfill

\end{document}